\documentclass{article} 
\usepackage{iclr2019_conference,times}

\usepackage{amsmath,amsfonts,bm}

\def\eqref#1{equation~\ref{#1}}

\def\1{\bm{1}}

\DeclareMathAlphabet{\mathsfit}{\encodingdefault}{\sfdefault}{m}{sl}
\SetMathAlphabet{\mathsfit}{bold}{\encodingdefault}{\sfdefault}{bx}{n}

\usepackage{xcolor}

\usepackage{hyperref}
\usepackage{url}
\usepackage{graphicx}

\title{Training on test data: \\
Removing near-duplicates in Fashion-MNIST }

\author{Christopher Geier \\
University of Virginia \\
School of Engineering and Applied Science \\ 
\texttt{cpg3rb@virginia.edu}
}

\iclrfinalcopy 
\begin{document}

\maketitle

\begin{abstract}
Fashion-MNIST is a popular machine learning benchmark task that improves on MNIST by introducing a harder problem, increasing the diversity of testing sets, and more accurately representing a modern computer vision task. In order to increase the data quality of Fashion-MNIST, this paper investigates near duplicate images between training and testing sets. Near-duplicates between testing and training sets artificially increase the testing accuracy of machine learning models. This paper identifies near-duplicate images in Fashion-MNIST and proposes a dataset with near-duplicates removed. 
\end{abstract}

\section{Introduction}

MNIST is a common benchmarking data set in the machine learning for the classification task of recognizing handwritten digits. The simplicity of the problem results in very high testing accuracy \citep{lecun-mnisthandwrittendigit-2010}. Modern machine learning approaches can easily achieve 97\% testing accuracy on MNIST. This high testing accuracy makes comparisons between machine learning algorithms less useful. Fashion-MNIST was proposed as a better alternative to MNIST to solve this problem \citep{xiao2017fashionmnist}. The Fashion-MNIST dataset consists of labeled clothing images retrieved from an online shopping site. Because the images more diverse, Fashion-MNIST makes the problem of distinguishing classes more difficult for algorithms. Fashion-MNIST hopes to more accurately give feedback on the performance of models. 

\begin{figure}[b]
\begin{center}
\includegraphics[width=.6\linewidth]{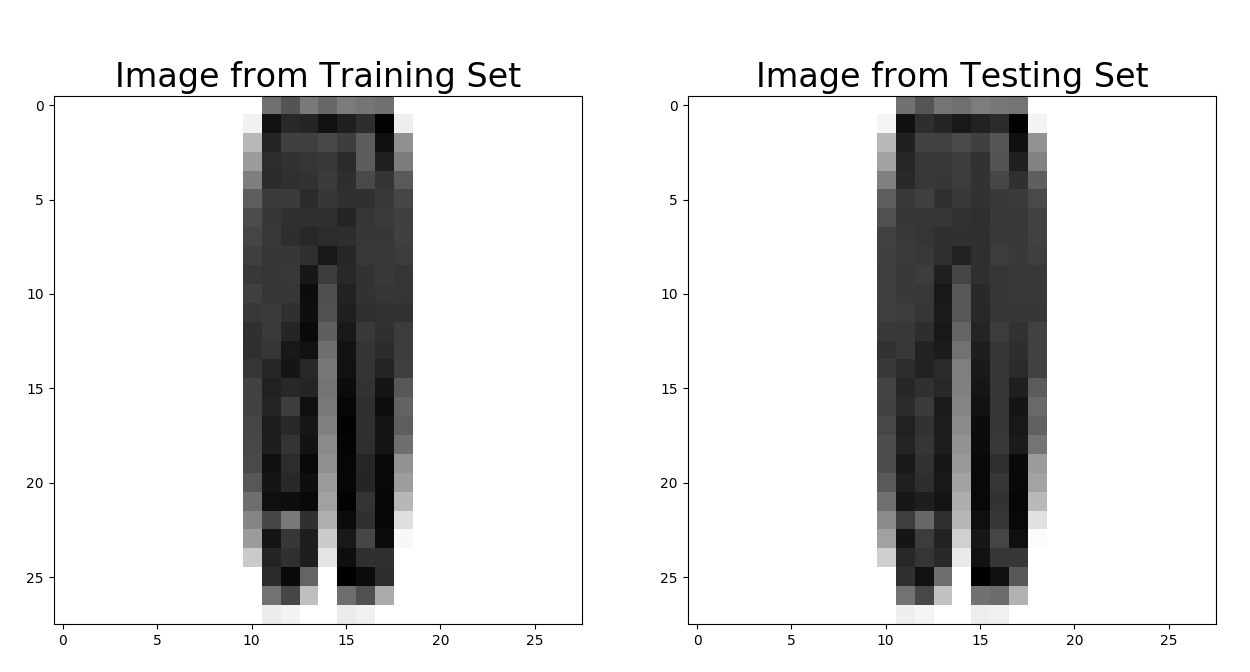}
\end{center}
\caption{Very similar images sample.}\label{fig:samples}
\end{figure}

Several authors have shown that popular machine learning datasets are often of poor quality in ways that can distort model evaluations. \citep{barz2019train} estimated that 3.3\% of images in the CIFAR-10 dataset are shared between the training and testing sets . \cite{radford2019language} showed that this problem goes beyond images. Popular natural language processing data-sets were found to have overlap of 1.63\% and 13.19\% between training and testing sets. By having samples shared between training and testing sets, models that over-fit the training set and memorize examples will also have an artificially high testing accuracy. A sample of very similar images found in Fashion-MNIST is shown in Figure~\ref{fig:samples}. In this paper, very similar images appearing both in the training and test of Fashion-MNIST are identified and removed to provide an improved test set.

\section{Sorting by similarity}
\label{headings}

To identify pairs of similar images, the pairs were ranked by similarity and then classified by a human analyst. Similarity of image pairs were calculated using feature vectors from a trained model. Using this measure of similarity, the image pairs were sorted. The most similar pairs were classified as very similar or different by a human analyst. Without using this method to rank similarity of image pairs, the number of image pairs to classify using a human analyst would have been impractical.

\subsection{Feature vectors}

As in \citet{barz2019train}, we use a convolutional neural network to identify similar images in the training and testing set. This was done by removing the last few layers of a trained neural network. At later layers convolutional neural networks identify higher level features present in images \citep{zeiler2013visualizing}. This can be used to identify images that have similar features, not just similar pixel values. We used a sequential neural network with three convolutional layers, batch normalization, max pooling and dropout. The target model architecture was referenced from \citet{googlefashionmnist2018colab}. After the feature vectors were extracted from the trained neural network, the distances between vectors were computed and ranked. Images that are similar have closer feature vectors, and this was used to rank the image pairs from most similar to most distinct.

\subsection{Identifying very similar versus distinct}

The process of selecting very similar images is very subjective task. The goal of this process is to remove images that the trained algorithm has already been trained on in the training set. Because algorithms being tested have been trained on very similar images, the likelihood that they will identify correctly should be increased. However there are issues with removing images from the testing set. The testing set is meant to represent the performance on a real-world problem. By removing images the distribution of real-world images is changed. 

This may be a problem for testing sets that are actually applied to real-world problems, however it is not as important for the testing accuracy on benchmarking sets like MNIST and Fashion-MNIST to mimic the real-world, but more important to show algorithms are learning features from the training dataset. Removing very similar images present in both the training and testing set may improve the intended purpose of these datasets to benchmark the generalization of algorithms beyond images in the training set. This also means that the very-similar image identification can be less conservative in identifying very similar images. 

The following characteristics were used to identify images that were considered "very similar":
\begin{itemize}
    \item Size and Shape: Overall figure must match by 10 pixels or less
    \item Outline: Aliasing around the clothing article is 90\% similar
    \item Tone: Difference of color is indistinguishable at a glance
    \item Features: Vary by 1 feature such as buttons, zippers, patterns, or text
\end{itemize}
\begin{figure}[tbh]
\begin{center}
\includegraphics[width=.6\linewidth]{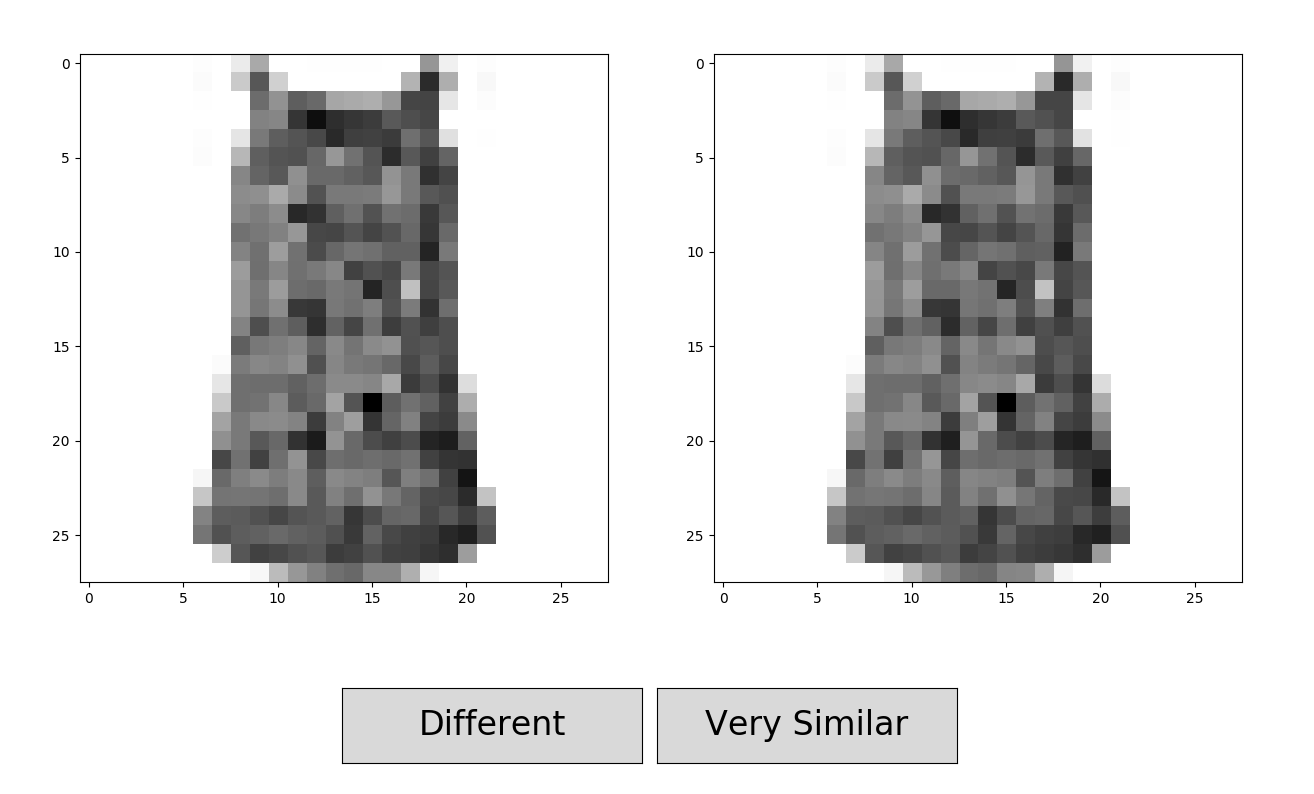}
\end{center}
\caption{User interface used to label images as distinct versus similar}\label{fig:ui}
\end{figure}

To collect the data, a human analyst can be used to mark images as very similar versus distinct. The image in Figure~\ref{fig:ui} showcases the user interface used to classify image pairs. Future work could involve improving this interface to allow for answers on a scale from distinct to identical.

Using this interface the analyst looked at pairs of images and determined their similarity using the characteristics above. Once the analyst identified 20 images in a class as different, a different class of images were presented. The images that were identified as being very similar were then removed from the testing set to create the new Fair-Fashion-MNIST dataset.

\begin{figure}[h]
\begin{center}
\includegraphics[width=.6\linewidth]{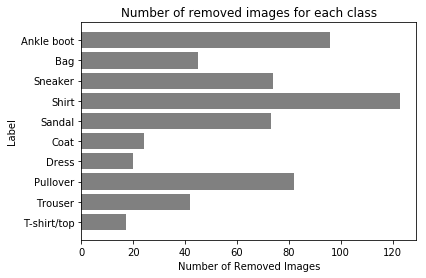}
\end{center}
\caption{Number of images removed from the testing set for each class}\label{fig:numremoved}
\end{figure}

\section{Results}
\label{others}

\begin{table}[tb]
\caption{Testing accuracy on benchmark datasets}\label{ref:comparison}
\begin{center}
\begin{tabular}{lccc}
\multicolumn{1}{c}{\bf Model}  &\multicolumn{1}{c}{\bf MNIST} &\multicolumn{1}{c}{\bf Fashion-MNIST} &\multicolumn{1}{c}{\bf Fair-Fashion-MNIST}
\\ \hline \\
SGD Classifier         &0.913 &0.829 &0.820\\
Perceptron             &0.845 &0.759 &0.754\\
Decision Tree          &0.877 &0.789 &0.783\\
Random Forest          &0.949 &0.844 &0.840\\
\end{tabular}
\end{center}
\end{table}

The number of images removed from the testing set for each class is shown in Figure~\ref{fig:numremoved}. In total, 5.98\% of the 10,000 original images in the testing set were removed. An accuracy comparison of trained algorithms is presented in Table~\ref{ref:comparison}. Although not as significant of an improvement as Fashion-MNIST made to MNIST, the near-similar removal process in this paper did reduced testing accuracy by a small amount. The Fair-Fashion-MNIST dataset and full benchmarking results can be found at \hyperlink{https://github.com/cpgeier/fair-fashion-mnist}{\sf\small https://github.com/cpgeier/fair-fashion-mnist}. The resulting dataset is compatible with existing MNIST and Fashion-MNIST scripts using the IDX file format. 

\section{Conclusion}

In this paper, a new dataset is introduced that limits the overlap of images between testing and training set to give more meaningful feedback to machine learning researchers. By removing overlapping images between testing and training sets, we produce a more fair testing set that should better predict the classifier's performance. The testing accuracy of common machine learning algorithms on our Fair-Fashion-MNIST dataset was slightly lower than on the original Fasion-MNIST. This likely shows an improvement in the testing set as the models will have an expected higher accuracy on images very similar to images in the training set. The resulting dataset is public and can be used for future Fashion-MNIST work. 

\section*{Acknowledgments}

I would like to thank my research advisor, David Evans, for guiding this research and helping me throughout conducting this research.

\bibliography{fair-fashion-mnist}
\bibliographystyle{iclr2019_conference}

\end{document}